\documentclass[lettersize,journal]{IEEEtran}
\usepackage{algorithmic}
\usepackage{array}
\usepackage[caption=false,font=normalsize,labelfont=sf,textfont=sf]{subfig}
\usepackage{textcomp}
\usepackage{stfloats}
\usepackage{verbatim}
\def\BibTeX{{\rm B\kern-.05em{\sc i\kern-.025em b}\kern-.08em
    T\kern-.1667em\lower.7ex\hbox{E}\kern-.125emX}}
\usepackage{balance}

\usepackage{cite}
\usepackage[utf8]{inputenc} 
\usepackage[T1]{fontenc}    
\usepackage{hyperref}       
\usepackage{url}            
\usepackage{booktabs}       
\usepackage{amsfonts}       
\usepackage{nicefrac}       
\usepackage{microtype}      
\usepackage[table]{xcolor}  
\usepackage{graphicx}
\usepackage{wrapfig}
\usepackage{amssymb}
\usepackage{amsmath, mathrsfs}
\usepackage{enumitem}
\usepackage[ruled, vlined, linesnumbered]{algorithm2e}
\usepackage{bm}
\usepackage{pifont}
\usepackage{booktabs}
\usepackage{threeparttable}
\usepackage{multicol}
\usepackage{dsfont}
\usepackage{tikz,tikz-qtree}
\usepackage{makecell}
\usepackage{multirow}

\usepackage{soul}
\usepackage[normalem]{ulem}
\renewcommand\sout{\bgroup\markoverwith
{\color{magenta}{\rule[.5ex]{2pt}{1pt}}}\ULon}

\usetikzlibrary{trees}

\newtheorem{definition}{Definition}[section]

\definecolor{color1}{RGB}{193,226,245}
\definecolor{color2}{RGB}{239,179,214}
\definecolor{color3}{RGB}{186,221,197}

\newcommand{\update}[1]{{\textcolor{black}{{#1}}}}  

\begin{document}
\title{A Survey on Safety-Critical Driving Scenario Generation -- A Methodological Perspective}

\author{Wenhao Ding, Chejian Xu, Mansur Arief, Haohong Lin, Bo Li, Ding Zhao
\IEEEcompsocitemizethanks{\IEEEcompsocthanksitem Wenhao Ding, Mansur Arief, Haohong Lin, and Ding Zhao were with the Department
of Mechanical Engineering, Carnegie Mellon University, Pittsburgh,
PA, USA. e-mail: \texttt{\{wenhaod, marief, haohongl\}@andrew.cmu.edu}, \texttt{dingzhao@cmu.edu}.\protect\\
\IEEEcompsocthanksitem Chejian Xu and Bo Li were with the Computer Science Department, University of Illinois Urbana-Champaign, Urbana, IL, USA. e-mail: \texttt{chejian2@illinois.edu}, \texttt{lbo@illinois.edu}.}
}

\markboth{Journal of \LaTeX\ Class Files}%
{A Survey on Safety-Critical Driving Scenario Generation -- A Methodological Perspective}

\maketitle

\begin{abstract}
Autonomous driving systems have witnessed significant development during the past years thanks to the advance in machine learning-enabled sensing and decision-making algorithms. One critical challenge for their massive deployment in the real world is their safety evaluation. Most existing driving systems are still trained and evaluated on naturalistic scenarios collected from daily life or heuristically-generated adversarial ones. However, the large population of cars, in general, leads to an extremely low collision rate, indicating that safety-critical scenarios are rare in the collected real-world data. Thus, methods to artificially generate scenarios become crucial to measure the risk and reduce the cost. In this survey, we focus on the algorithms of safety-critical scenario generation in autonomous driving. We first provide a comprehensive taxonomy of existing algorithms by dividing them into three categories: data-driven generation, adversarial generation, and knowledge-based generation. Then, we discuss useful tools for scenario generation, including simulation platforms and packages. Finally, we extend our discussion to five main challenges of current works -- fidelity, efficiency, diversity, transferability, controllability -- and research opportunities lighted up by these challenges.
\end{abstract}

\begin{IEEEkeywords}
Autonomous vehicles, Safety, Robustness, Deep Generative Models
\end{IEEEkeywords}

\section{Introduction}
\label{sec:introduction}

\IEEEPARstart{A}{rtificial} Intelligence (AI) has been widely used in software products such as facial recognition~\cite{masi2018deep} and voice-print verification~\cite{saquib2010survey}. But as AI continues to grow and research has expanded to physical products like autonomous vehicles (AVs), the question of safety is now at the forefront of this cutting-edge field.
The reason why intelligent physical systems are much harder to be deployed is that our world is complicated and long-tailed, causing too much uncertainty to the intelligent agents. 
The driving skills would take several months to learn, even for us humans, due to the complex traffic scenarios.
Therefore, the AVs should be trained and evaluated on lots of different scenarios to demonstrate their safety and capability of dealing with diverse situations~\cite{zhong2021survey, riedmaier2020survey}. 

\begin{figure*}
    \centering
    \includegraphics[width=0.95\textwidth]{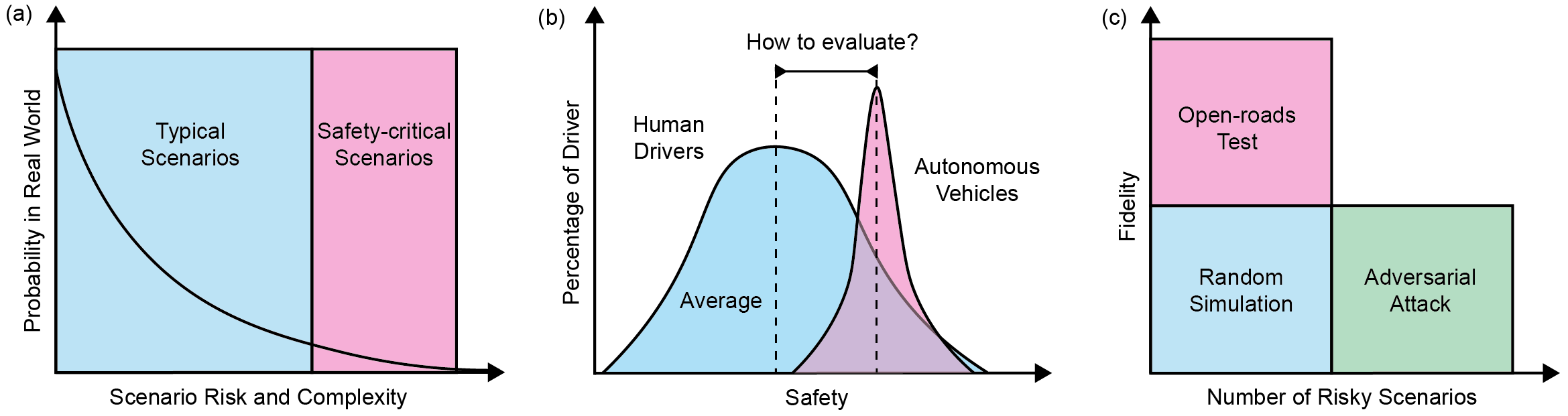}
    \caption{The overview of autonomous vehicle evaluation. (a) Most of the scenarios that happen in the real world are typical scenarios; safety-critical scenarios are extremely rare. (b) AVs are supposed to have higher average safety than human drivers, but the gap is not easy to be evaluated and measured. (c) Comparison between different generation methods. Most existing methods cannot satisfy the fidelity and safety-critic metrics simultaneously.}
    \label{fig:overview}
\end{figure*}

According to the 2020 disengagement report from the California Department of Motor Vehicle~\cite{disengagement}, there were at least five companies (Waymo, Cruise, AutoX, Pony.AI, Argo.AI) that made their AVs drive more than 10,000 miles without disengagement. These results are usually recorded in normal driving scenarios without risky situations. 
It is a great achievement that current AVs are successful in normal cases trained by hundreds of millions of miles of training. 
However, we are still not sure whether AVs have enough safety and robustness in distinct scenarios. 
For example, when one AV is driving on the road, a kid suddenly runs into the drive lane chasing a ball. 
This emergency case leaves the AV a very short time to react, and even a subtle misbehave could cause vital damage. 
This kind of situation is named \textit{safety-critical} scenarios and is usually extremely rare in the normal driving case, as shown in Figure~\ref{fig:overview}. 

To efficiently evaluate the safety of AVs, more and more people from the government, industry, and academia start to focus on the generation of safety-critical scenarios.
National Highway Traffic Safety Administration (NHTSA), a government agency of the United States Department of Transportation, summarized driving system testable cases~\cite{thorn2018framework} and pre-crash scenarios~\cite{najm2007pre} as well as published a review of simulation frameworks and standards related to driving scenarios~\cite{schnelle2019review}.
Waymo, one of the leading companies in autonomous driving, released a safety report to illustrate how they reconstruct fatal crashes in simulations from collected data~\cite{scanlon2021waymo}. 
Meanwhile, academic research aims to generate safety-critical scenarios~\cite{eykholt2018robust, ding2020learning} surges by developing methods in Deep Generative Models (DGMs)~\cite{harshvardhan2020comprehensive} and adversarial attack techniques~\cite{goodfellow2014explaining}.
With the upcoming massive deployment of self-driving cars, an overview that systematically summarizes existing works in safety-critical scenario generation is urgently demanded.

The \textbf{objective of this survey} is to thoroughly review the literature on safety-critical scenario generation from a methodological perspective and provide a panorama of the approaches developed so far. The key contributions are as follows:
\begin{itemize}
    \item We built a taxonomy to categorize existing algorithms of safety-critical scenarios generation according to the information and general framework they leverage.
    \item We summarized the tools used for scenario generation, including autonomous driving simulators and open-sourced packages for scenario design, as well as commonly used scenario datasets.
    \item We identify five challenges of safety-critical scenario generation and corresponding future directions to further push the frontier. 
\end{itemize}

\noindent \textbf{Existing surveys.}
There have existed surveys~\cite{zhong2021survey, riedmaier2020survey, menzel2018scenarios} summarizing the scenario-based method for autonomous vehicle evaluation. 
Among these previous works, \cite{riedmaier2020survey} gives a clear and inspiring definition and categorization of traffic scenarios. 
Their framework divide the scenario generation problem into two processes: (1) scenario collection, where the scenarios either come from expert abstract knowledge or real-world collected dataset; (2) scenario selection, where the scenario is selected by sampling or feedback optimization. 
A recent survey~\cite{zhong2021survey} about scenario-based testing algorithms for autonomous driving systems divides a scenario into five layers (road level, traffic infrastructure, manipulation of previous two layers, object, environment) and discusses the parameters searching in each layer separately.
In \cite{menzel2018scenarios}, the authors propose another perspective by categorizing traffic scenarios into three layers: functional, logical, and concrete. This is a hierarchical structure since the abstracting level decreases from left to right, and the number of scenarios increases at the same time. These categorizations provide inspiring yet different perspectives on scenario-based testing. 
These works provide a great overview of the entire framework of the evaluation of autonomous driving systems.
We want to emphasize that our work differs from them in that we focus on the algorithms of \textit{safety-critical} scenario generation - the core component in evaluating safety and robustness. 
Our taxonomy is built according to the information that the generation algorithms leverage, focusing on how these generation methods deal with the structure of traffic participants in a scenario and how the autonomous vehicle interacts with surrounding objects. 
Moreover, we identify five challenges that limit the current generation method, which we hope would inspire future directions.

\noindent \textbf{Organization of this survey.} 
Section~\ref{sec:scenario} gives an overview of safety-critical scenarios, including the definition, representation, and metrics used for selecting the scenarios. 
Then, we divide the generation methods into three types: \textit{Data-driven Generation} (Section~\ref{sec:data}) purely samples collect dataset and uses density estimation models for generation; \textit{Adversarial Generation} (Section~\ref{sec:adversarial}) considers the ego vehicle during the generation and builds an adversarial learning framework. \textit{Knowledge-based Generation} (Section~\ref{sec:knowledge}) either uses pre-defined rules by experts or integrates external knowledge during the generation.
We also summarize the traffic simulators, datasets, and open-sourced packages that can accelerate the development of driving scenarios in Section~\ref{sec:tools}.
In Section~\ref{sec:directions}, we list five challenges of safety-critical scenario generation and discuss the potential research directions lighted up by these challenges.
Section~\ref{sec:conclusion} closes this survey with general conclusions and lessons learned from the current stage.

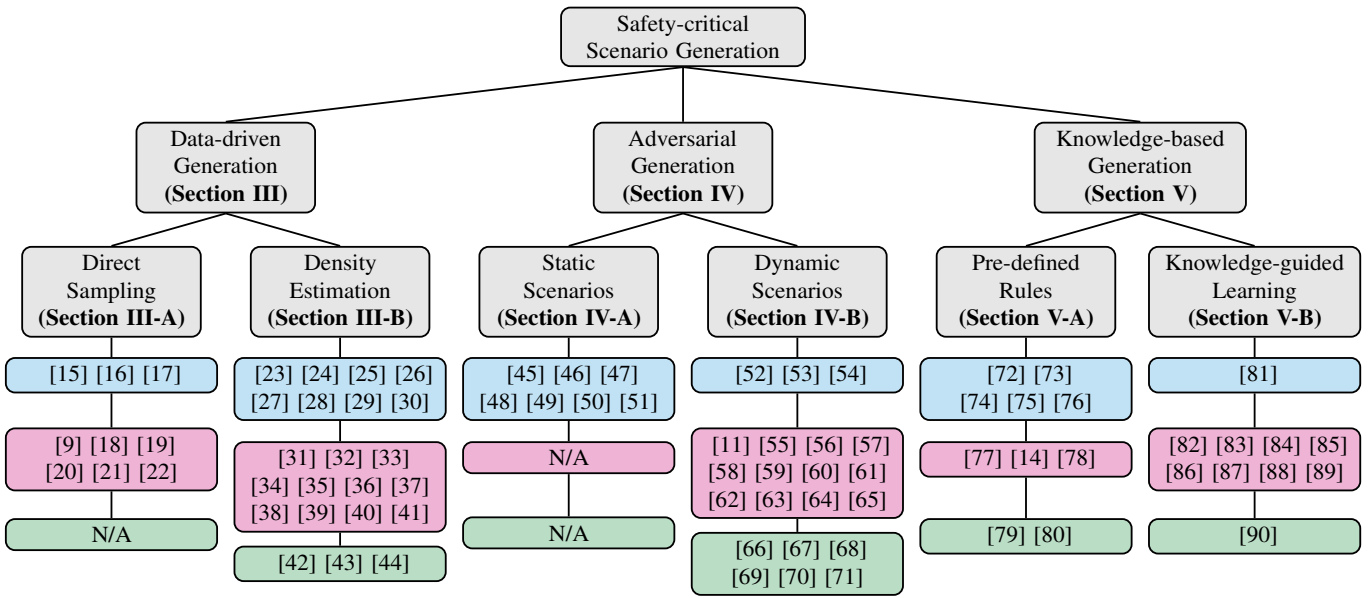
\begin{figure*}[t]
    \centering
    \resizebox{1.0\textwidth}{!}{\begin{tikzpicture}[
    l0/.style = {fill=gray!20, draw=black!100,thick, rounded corners, text width=30mm, align=center, anchor=north},
    l1/.style = {fill=gray!20, draw=black!100,thick, rounded corners, text width=25mm, align=center, anchor=north},
    l2/.style = {fill=gray!20, draw=black!100,thick, rounded corners, text width=30mm, align=center, anchor=north},
    l3/.style = {fill=gray!20, draw=black!100,thick, rounded corners, text width=35mm, align=center, anchor=north},
    perception/.style = {fill=color1, draw=black!100,thick, rounded corners, text width=30mm, align=center, anchor=north},
    planning/.style = {fill=color2, draw=black!100,thick, rounded corners, text width=30mm, align=center, anchor=north},
    control/.style = {fill=color3, draw=black!100,thick, rounded corners, text width=30mm, align=center, anchor=north},
    sibling distance = 42mm,
    ]
    
    \tikzset{edge from parent/.append style={thick}}
    \tikzstyle{level 1}=[sibling distance=70mm, level distance=13mm] 
    \tikzstyle{level 2}=[sibling distance=35mm, level distance=12mm] 
    \tikzstyle{level 3}=[sibling distance=35mm, level distance=10mm] 
    \tikzstyle{level 4}=[sibling distance=20mm, level distance=8mm] 
    \tikzstyle{level 5}=[sibling distance=35mm, level distance=9mm] 
    
    \node[l3] {Safety-critical \\ Scenario Generation}
        child{ 
            node[l1] {Data-driven \\ Generation \\ \textbf{(Section~\ref{sec:data})}}
            child{ 
                node[l1] {Direct \\ Sampling \\ \textbf{(Section~\ref{sec:data1})}}
                child{ 
                    node[perception] {
                    \cite{van2015automated}
                    \cite{fang2020augmented}
                    \cite{manivasagam2020lidarsim}}
                    child{
                        node[planning] {
                        \cite{scanlon2021waymo}
                        \cite{arief2018accelerated}
                        \cite{knies2020data}
                        \cite{kruber2018unsupervised}
                        \cite{kruber2019unsupervised}
                        \cite{wang2018extracting}}
                        child{
                            node[control] {N/A}
                        }
                    }
                }
            }
            child{ 
                node[l1] {Density \\ Estimation\\ \textbf{(Section~\ref{sec:data2})}}
                child{node[perception] {
                    \cite{yang2020surfelgan}  
                    \cite{chen2021geosim} 
                    \cite{ehrhardt2020relate}
                    \cite{li2019aads}
                    \cite{xiao2021synlidar}
                    \cite{kar2019meta}
                    \cite{devaranjan2020meta}
                    \cite{savkin2021unsupervised}}
                    child{
                        node[planning] {
                        \cite{ding2018new}
                        \cite{haakansson2021driving}
                        \cite{ding2020cmts}
                        \cite{huang2018synthesis}
                        \cite{guo2019modeling}
                        \cite{tan2021scenegen}
                        \cite{wen2020scenario}
                        \cite{wheeler2015initial}
                        \cite{wheeler2016factor}
                        \cite{wheeler2019critical}
                        \cite{wulfe2018real}}
                        child{
                            node[control] {
                            \cite{bojarski2016end}
                            \cite{ho2016generative}
                            \cite{chen2019deep}}
                        }
                    }
                }
            }
        }
        child{ node[l1] {Adversarial \\ Generation \\ \textbf{(Section~\ref{sec:adversarial})}}
            child{ node[l1] {Static \\ Scenarios \\ \textbf{(Section~\ref{sec:adversarial1})}}
                child{ node[perception] {
                    \cite{jain2019analyzing}
                    \cite{tu2020physically} 
                    \cite{abdelfattah2021towards} 
                    \cite{cao2019adversarial} 
                    \cite{prakash2019structured}
                    \cite{khirodkar2018vadra}
                    \cite{zhu2021adversarial}}
                    child{
                        node[planning] {N/A}
                        child{
                            node[control] {N/A}
                        }
                    }
                }
            }
            child{ node[l1] {Dynamic \\ Scenarios \\ \textbf{(Section~\ref{sec:adversarial2})}}
                child{ node[perception] {
                    \cite{ruiz2018learning}
                    \cite{o2018scalable} 
                    \cite{abeysirigoonawardena2019generating}}
                    child{
                        node[planning] {
                        \cite{ding2020learning} 
                        \cite{ding2021multimodal}
                        \cite{ghodsi2021generating}
                        \cite{wang2021advsim} 
                        \cite{nonnengart2019crisgen}
                        \cite{feng2021intelligent}
                        \cite{sun2021corner}
                        \cite{koren2018adaptive} 
                        \cite{koren2019efficient} 
                        \cite{koren2020adaptive} 
                        \cite{koren2021finding}
                        \cite{rempe2022generating}}
                        child{
                            node[control] {
                            \cite{arief2021deep}
                            \cite{arief2021certifiable}
                            \cite{chen2021adversarial}
                            \cite{wachi2019failure}
                            \cite{kuutti2020training} 
                            \cite{lee2015adaptive} }
                        }
                    }
                }
            }
        }
        child{ node[l0] {Knowledge-based \\ Generation \\ \textbf{(Section~\ref{sec:knowledge})}}
            child{ node[l1] {Pre-defined \\ Rules \\ \textbf{(Section~\ref{sec:knowledge1})}}
                child{ node[perception] {
                    \cite{cai2020summit}
                    \cite{carlachallenge}
                    \cite{li2021metadrive}
                    \cite{fremont2022scenic}
                    \cite{fremont2020formal}}
                    child{
                        node[planning] {
                        \cite{mcduff2021causalcity}
                        \cite{menzel2018scenarios}
                        \cite{bagschik2018ontology}
                        }
                        child{
                            node[control] {
                            \cite{rana2021building}
                            \cite{zhou2020smarts}}
                        }
                    }
                }
            }
            child{ node[l0] {Knowledge-guided \\ Learning \\ \textbf{(Section~\ref{sec:knowledge2})}}
                child{ node[perception] {
                    \cite{ding2021semantically}}
                    child{
                        node[planning] {
                        \cite{ding2021causalaf}
                        \cite{zhang2022adversarial}
                        \cite{cao2022robust}
                        \cite{klischat2019generating}
                        \cite{althoff2018automatic}
                        \cite{rocklage2017automated}
                        \cite{klischat2020scenario}
                        \cite{wang2021commonroad}}
                        child{
                            node[control] {
                            \cite{shiroshita2020behaviorally}}
                        }
                    }
                }
            }
        };
\end{tikzpicture}}
    \caption{Taxonomy of safety-critical scenarios generation methods. The colors of boxes denote the modules of AV system that the generation algorithms target on: \color{color1}{\textbf{Perception}}, \color{color2}{\textbf{Planning}}, \color{color3}{\textbf{Control}}.}
    \label{fig:taxonomy}
\end{figure*}

\section{Overview of Safety-critical Scenario}
\label{sec:scenario}

\subsection{Definitions}

In this survey, we define a scenario with static and dynamic contents in it and also consider the behavior of dynamic objects. Formally, we have the following definition:
\begin{definition}[Driving Scenario]
\textit{
The driving scenario is defined by a combination of three sets: $x \in \mathcal{X} = \{\mathcal{S}, \mathcal{I}, \mathcal{B} \}$. $\mathcal{S}$ represents the static environment, \update{including road shape, traffic sign, traffic light, etc.} $\mathcal{I}$ represents the initial condition and properties of dynamic objects. $\mathcal{B}$ represents the sequential behaviors of dynamic objects.}
\end{definition}

Separating the static and dynamic contents benefits the representation and generation of scenarios and has shown great success in~\cite{team2021open}. The road geometry and static traffic objects (e.g., traffic signs and traffic lights) determine the background of the scenario and the long-term goal of the testing. The dynamic objects that participate in the traffic enable complex interaction with the AV and influence its short-term decisions of it. 
\update{Note that in this definition, we do not have any specific requirements for the AV. Since modern AV algorithms are usually divided into several important components, we consider the following definition of an AV system:}
\begin{definition}[AV System]
\textit{\update{
An AV system is denoted as $F(x)$, which consists of three modules: $F(x) = F_{per} \circ F_{pla} \circ F_{con}$. $F_{per}$ denotes the perception module that takes sensor data in and outputs locations of surrounding objects. $F_{pla}$ denotes the planning module that provides a feasible trajectory using the prediction of potential obstacles. $F_{con}$ denotes the control module that executes control signals to follow given waypoints.}}
\end{definition}

With the above two definitions, we then define the objective of safety-critical scenario generation:
\begin{definition}[Safety-critical Scenario Generation]
\textit{Assume the distribution of scenario $x$ is parametrized by $\theta$, the safety-critical scenarios can be generated by}
\begin{equation}
    \hat{\theta} = \arg\max \mathbb{E}_{x\sim p_{\theta}(x)}\left[g(F(x)) \right]
\end{equation}
\textit{\update{where $g(\cdot)$ is the metric function that measures the risk level and safety-related properties (e.g., collision rate and distance
driven out of road). $F(x)$ is the AV system that takes scenario $x$ in and outputs a sequence of control signals.}}
\label{def:scenario_generation}
\end{definition}
This optimization problem is not easy to solve mainly because of two aspects: a) the representation of the distribution $p_{\theta}(x)$, and b) the selection of the metric $g(\cdot)$. We will discuss how existing methods deal with these two problems in the following two subsections, respectively.

\subsection{Representation of Scenario}

We have multiple choices to represent the scenario, and generally, the selection depends on the module to be evaluated. 
If the target module is a perception module, we use high-dimensional sensing data, such as images and LiDAR point cloud. Directly generating high-fidelity sensing data is quite difficult, and there are an increasing number of works focused on this area~\cite{karras2020analyzing}. An alternating method is leveraging differential renderer~\cite{kato2018neural, liu2019soft} and LiDAR simulator~\cite{ding2021semantically, tu2020physically} to generate high-dimensional data with ray casting algorithms. 
If we want to evaluate motion planning or control module, we can turn to lower dimensions. We can use either trajectories or policy models of dynamic objects. Using a trajectory is less flexible than using a policy model, but the scenarios with trajectory representation are more controllable. 

\subsection{Metrics for Generation}

\update{
The most important factor of solving the optimization problem Definition~\ref{def:scenario_generation} is the metric of risk $g(\cdot)$. 
A proper risk metric makes the generated scenarios useful for discovering failure cases of AV systems, while an improper risk metric only provides useless scenarios that either is too trivial for AV systems or too rare to happen in the real world.
}
The level of risk is mainly reflected by the interaction between the autonomous vehicle and participants in the scenario, which can be naturally described by the distance -- a small distance means the risk of collision is high. This intuition can be described by Time-to-Collision (TTC)~\cite{hayward1972near}:
\begin{equation}
    TTC_{F}(t) = \frac{X_{L}(t) - X_{F}(t) - l_{L}}{\dot{X}_{F}(t) - \dot{X}_{L}(t)},\ \forall \ \dot{X}_{F}(t) > \dot{X}_{L}(t),
\end{equation}
where $X$ denotes the position, $\dot{X}$ denotes the derivative of $X$ with respect to time or the speed, and $l_L$ denotes the leading vehicle's length; $L$ and $F$ as subscripts refer to leading and following vehicles in a car-following process.
\update{Following this time-based metric, there are numerous variants of TTC such as Time Exposed TTC (TET)~\cite{minderhoud2001extended} and Modified TTC (MTTC)~\cite{ozbay2008derivation}. 
}

\update{
Another two main types of metrics are distance-based and deceleration-based metrics. The first one uses the distance available to avoid a collision, e.g., Proportion of Stopping Distance (PSD)~\cite{allen1978analysis}. The second one defines dangerous situations using the rate deceleration during an emergency, e.g., Deceleration Rate to Avoid the Crash (DRAC)~\cite{almqvist1991use}.
Please check~\cite{mahmud2017application} for more metrics belonging to these two types.
In addition, we can also use the post-hoc analysis of unsafe behavior to indicate the risk, for example, collision rate, average distance driven out of the road, and the frequency of running stop signs and red lights~\cite{carlachallenge, li2021metadrive}.
}


However, measuring only the risk is not enough for scenario generation. We need to ensure that the scenarios are realistic and able to happen in the physical world. We also want to discover as many as possible scenarios rather than a single one. In fact, these kinds of constraints make the generation of scenarios a much harder task.
We will discuss this topic more in Section~\ref{sec:directions}.

Finally, we highlight the need to balance the effort in modeling the input scenario with the effort in utilizing the scenario to compute the final metrics. This is of particular importance since the uncertainty in the final metrics is due to both modeling uncertainty and sampling uncertainty. While the former can be driven down by designing as accurate a model as possible, the latter can only be reduced by running longer, and larger number of simulation runs \cite{huang2019evaluation}. A good review of this input-induced uncertainty and ways to deal with it in modeling can be found in \cite{henderson2003input, barton2010framework, shafaei2018uncertainty}.

\begin{figure*}
    \centering
    \includegraphics[width=1.0\textwidth]{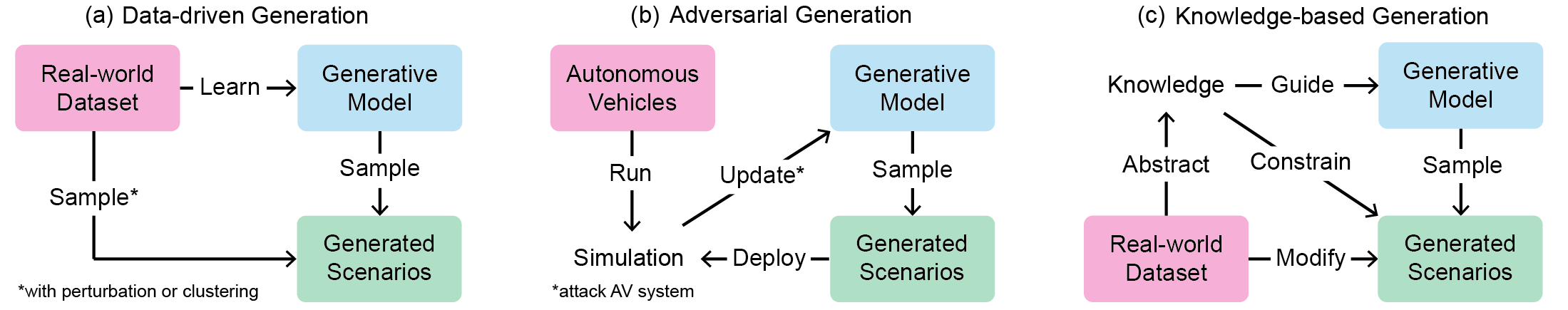}
    \caption{Illustration of three types of generation methods. (a) Data-driven methods only use the collected data to sample directly or via generative models. (b) Adversarial methods use the feedback from the autonomous vehicle that is deployed in the simulation. (c) Knowledge-based methods leverage the information mainly from external knowledge as constraints or guidance to the generation.}
    \label{fig:three_generation}
\end{figure*}

\section{Data-driven Generation}
\label{sec:data}

In this section, we consider the algorithms that only leverage information from the collected datasets. 
\update{For example, we control several vehicles to pass an intersection by making them follow the trajectories recorded in the same intersection in the real world.}
These methods are mainly divided into two parts. The first part directly samples from the dataset $x \sim \mathcal{D}$ to reproduce the real-world log, which usually suffers from the problem of rareness.
The second part is using density estimation models (e.g. DGMs) $p_{\theta}(x)$ parameterized by $\theta$ to learn the distribution of scenarios, which enables the generation of unseen scenarios. Usually, the learning objective of these models is maximizing the log-likelihood
\begin{equation}
    \hat{\theta} = \arg\max \sum_{x} \log p_{\theta}(x),
\end{equation}
and the sampling process is conducted by $x \sim p_{\theta}(x)$ from random noises.

\subsection{Direct Sampling}
\label{sec:data1}

An intuitive way of generation is directly sampling from a collected dataset which reproduces the scenario from the road test log. According to different algorithms used before and during the sampling, we summarize the following three groups.

\subsubsection{Data Replay}
Most autonomous driving companies maintain scenario bases to store the scenarios they recognized as important~\cite{webb2020waymo}.
During this process, one crucial step is the tools that can automatically convert the scenario from log to virtual simulations~\cite{van2015automated}.
In addition, efficiently selecting the critical scenario from a huge number of scenarios is another problem.
\cite{knies2020data} extracts useful scenarios according to the cooperative actions for cooperative maneuver planning evaluation.
In \cite{arief2018accelerated}, the authors developed a method to select a testing ground to accelerate the performance estimation of AVs performance on public streets, where the main contribution is describing the risk intensities of the traffic system in an area of interest with Non-Homogeneous Poisson Process model~\cite{pham2003nhpp}.

\subsubsection{Clustering}
To better categorize the collected scenario, \cite{kruber2018unsupervised} and \cite{kruber2019unsupervised} propose to use unsupervised clustering methods to put similar scenarios into groups, which helps improve the efficiency of testing AVs on a specific type of scenario.
However, clustering the entire scenario might be inefficient and inaccurate since the scenarios are usually complex and composed of finite building blocks. The concept of \textit{Traffic Primitive} is proposed in~\cite{wang2018extracting} to represent those blocks. They use the Hierarchical Dirichlet Process Hidden Markov Model (HDP-HMM)~\cite{fox2008hdp}, a non-parametric Bayesian learning method, to unsupervised extract primitives from scenarios to better cluster similar scenarios.

\subsubsection{Random Perturbation}
The main obstacle to direct conversion is that the diversity of scenarios is limited. 
Therefore, recent works from some leading companies have started to use random perturbation to augment the number of scenarios.
Baidu creates a physical-based LiDAR model~\cite{fang2020augmented} to transfer a dense point cloud to match the line-style output of the LiDAR sensor. By randomly placing the Computer-Aided Design model of vehicles and pedestrians, their algorithms can generate a huge number of scenarios. 
Similarly, Uber builds a more precise LiDAR model~\cite{manivasagam2020lidarsim} using neural networks (NNs) to mimic the reflection details of the real-world sensors.
Besides the high-dimensional representation, Waymo tries to reconstruct more fatal crashes from collected data by randomly perturbing important parameters~\cite{scanlon2021waymo}. 

\subsection{Density Estimation Methods}
\label{sec:data2}
Consider the driving scenarios following a distribution, and then we can use collected data to learn a density model to approximate this distribution. We divide this type of algorithm into three categories according to the density model they use.

\subsubsection{Bayesian Networks}

Bayesian Networks is a probabilistic graphical model that uses nodes to represent objects and edges to represent the relation between nodes. This structured model can naturally describe the objects in the scenario.
\cite{wheeler2015initial} uses Dynamic Bayesian Network~\cite{ghahramani1997learning} to model the complex traffic scenarios, and \cite{wheeler2016factor} uses factor graphs to model driving behaviors between multiple vehicles.
In \cite{wulfe2018real}, Importance Sampling (IS) is used to sample driving scenarios represented by Bayesian Networks.
In \cite{wheeler2019critical}, the authors try to find as many risky scenarios as possible and cluster them. They first uniformly sample from the clusters and then use IS to sample the specific scenarios represented by factor graphs.
Besides using Dynamic Bayesian Network or factor graphs to model the structures of traffic participants, Gaussian Process (GP) is a powerful non-parametric method to model the distribution of sequential scenarios~\cite{huang2018synthesis}.
Following this direction, \cite{guo2019modeling} combines GP and Dirichlet Process to build a model with an infinite number of clusters to discover traffic primitives~\cite{wang2018extracting}, which can be combined to create new scenarios.
\cite{xiao2021synlidar} generates point cloud sequential datasets via minimizing the gap between real-world LiDAR and simulation data.
Similarly, Meta-sim~\cite{kar2019meta} and Meta-Sim2~\cite{devaranjan2020meta} try to minimize the sim-to-real gap to reconstruct traffic scenarios for automatic labeling.
Their main contribution is that they use a scene graph to represent the scenario, which is a hierarchical structure that makes the generation more efficient.
In \cite{savkin2021unsupervised}, the scene graph is also used to generate image scenarios.

\subsubsection{Deep Learning Models}

Deep learning models are also introduced into the generation in \cite{tan2021scenegen} and \cite{wen2020scenario}.
\cite{tan2021scenegen} inputs the current state of the AV and a high-definition map to a Long Short Term Memory (LSTM)~\cite{hochreiter1997long} module to sequentially generate the trajectory of surrounding vehicles and pedestrians.
They train their model with normal traffic data since their target is to generate naturalistic scenarios.
\cite{wen2020scenario} proposes a quite complex system to generate scenarios in a simulator, which uses Convolution Neural Network (CNN)~\cite{o2015introduction} as a selector to generate agents surrounding the AV.
In TrafficSim~\cite{suo2021trafficsim}, both Gated Recurrent Unit (GRU) and CNN are used to learn the behaviors of multi-agents from real-world data. This method can generate realistic multi-agent traffic scenarios.

\subsubsection{Deep Generative Models (DGMs)}
Recently, DGMs have shown great success in generating image and voice data. 
There are five types of modern generative models: 
Generative Adversarial Nets (GAN)~\cite{goodfellow2014generative}, Variational Auto-encoder (VAE)~\cite{kingma2013auto}, Autoregressive Models~\cite{van2016pixel, oord2016wavenet}, Flow-based model~\cite{dinh2016density, chen2018neural}, Diffusion model~\cite{ho2020denoising, song2020denoising}.
The readers can find more details about DGMs in this survey~\cite{harshvardhan2020comprehensive}.
With the power of VAE, \cite{ding2018new} learns a latent space of encounter trajectories and generates unseen scenarios by sampling from the latent space. 
However, with less understanding of the latent code, the generation is not controllable.
In \cite{ding2020cmts}, the authors propose CMTS, which combines normal and collision trajectories to generate safety-critical scenarios by doing interpolation in the latent space.
As for the usage of GAN, \cite{haakansson2021driving} introduces recurrent models to generate realistic scenarios of highway lane changes. They use real-world data in the discriminator to help the improvement of the generator.
The advantage of DGMs is that they can learn a low-dimensional latent space of high-dimensional and structured data using NNs. 
Therefore, we can easily generate high-dimensional sensing scenarios. 
SurfelGAN is proposed in \cite{yang2020surfelgan} to directly generate point cloud data to represent scenarios from the view of the AV.
\cite{chen2021geosim} can add new vehicles to collected driving videos to generate realistic video scenarios, where they also consider the motion planning of vehicles.
\cite{ehrhardt2020relate} generates traffic videos with multi-object scene synthesis using a GAN framework. To make the video realistic, they integrate physical conditions into the generation.
A data-driven scenario simulator is designed in \cite{li2019aads} to generate both LiDAR and trajectory data to augment the diversity of driving scenarios.

\update{\subsubsection{Imitation learning}
Another way of mimicking the dataset is using imitation learning (IL) methods~\cite{hussein2017imitation}, which takes the observation as input and directly outputs the behavior to control agents. The training of IL is exactly the same as supervised learning. After training, the IL model~\cite{bojarski2016end, ho2016generative, chen2019deep} can reproduce the same behavior as real-world agents when the model encounter the same observation. However, when the observation is not covered by the dataset, the behavior of the model could be unreasonable and unpredictable. 
}

\section{Adversarial Generation}
\label{sec:adversarial}

\update{In this section, we consider a more efficient way for generation, which actively creates risky scenarios by attacking the AV system.
For example, we control a pedestrian to cross the road and intentionally make it collide with the AV. Although the AV may avoid the collision in most cases, we can still obtain safety-critical scenarios where accidents happen.}
This framework, named adversarial generation, consists of two components, one is the generator, and the other is the victim model, i.e., the AV. Then the targeted generation process can be formulated as
\begin{equation}
    \hat{\theta} = \arg\min \mathbb{E}_{p_{\theta}(x)} [Q(x, \pi)]
\end{equation}
where $Q(\cdot, \pi)$ is a quantitative function to indicate the performance of the policy $\pi$ taken by the AV. 
We notice that the adversarial generation will mainly focus on a specific small set of scenarios therefore it would be good to consider the diversity by adding constraint or entropy of the distribution $\mathbb{H}(x)$.
Since we consider the influence of AV, this type is also named Vehicle-in-the-loop testing in previous work~\cite{fernandez2021trustworthy}.
Since the autonomous driving system consists of multiple modules, we divide these methods according to the type of victim models. When the model is used for single frame inputs, e.g., object detection and segmentation, we only need to generate static scenarios. When the victim model requires a sequential testing case, we generate dynamic scenarios which contain the motion of all objects.

\subsection{Static Scenarios Generation}
\label{sec:adversarial1}

Under the adversarial framework, directly generating high-dimensional data could be challenging. Thus most methods generate the poses of objects and then use renders to get the final output.
\cite{jain2019analyzing} learns the poses of vehicles and uses a differentiable render to get the first-point-view images to attack object detection algorithms.
\cite{prakash2019structured} and \cite{khirodkar2018vadra} extend the original {Domain Randomization} methods~\cite{tobin2017domain} to the adversarial generation.
\cite{prakash2019structured} proposes {Structured Domain Randomization}, which uses a Bayesian Network to actively generate the poses of vehicles.
\cite{khirodkar2018vadra} proposes {Adversarial Domain Randomization}, which targets the scenario of parking lots.

All of the above methods focus on image generation, but some works generate point cloud scenarios. 
\cite{tu2020physically} puts an adversarial object on the top of a vehicle and optimizes the shape of the object to make the vehicle disappear in LiDAR detection algorithms.
Similarly, \cite{abdelfattah2021towards} shares the same idea but uses both LiDAR and image information.
After the optimization of the shape of the object, \cite{cao2019adversarial} use 3D printing to build the object in the real world. 
The experiment shows that the object on the road is indeed ignored by the detection algorithms.
In addition to using renders, \cite{zhu2021adversarial} tries to directly add new points to the existing point cloud to attack segmentation algorithms.

\subsection{Dynamic Scenario Generation}
\label{sec:adversarial2}

When the target is to evaluate planning and control modules, the scenarios are required to be dynamic and sequential. 
For these kinds of algorithms, we further divide current works into two types according to the flexibility of the scenario.

\subsubsection{Initial Condition}
The first type is controlling the initial conditions of the scenario (e.g., initial velocity and spawn position) or providing the entire trajectory at the beginning. The advantage is the low dimension of search space and the little computational resource required.
In \cite{ding2020learning} and \cite{ding2021multimodal}, the authors generate the initial poses of a cyclist to attack the AV. \cite{ding2021multimodal} approximates the parameters with a Gaussian distribution, which limits the diversity of generated scenario. 
As a remedy, \cite{ding2021multimodal} use normalizing flow models to learn a multi-modal distribution. It also uses the distribution of real-world data as a constraint to improve fidelity.
Parallel work~\cite{ghodsi2021generating} generates a set of possible driving paths and identifies all the possible safe driving trajectories that can be taken starting at different times.
Similarly, \cite{wang2021advsim} directly optimizes existing trajectories to perturb the driving paths of surrounding vehicles. They use Bayesian Optimization~\cite{frazier2018tutorial} for the optimization, and the scenario is represented with point cloud.
\update{To generate real-world traffic scenarios, \cite{rempe2022generating} optimizes the adversarial trajectory in the latent space of a VAE model.}
Besides controlling the dynamic objects, some works also focus on searching the weather parameters (e.g., sun and rain) to create different scenarios.
\cite{ruiz2018learning} searches the type of weather using REINFORCE~\cite{williams1992simple} algorithm.
\cite{o2018scalable} uses {Generative Adversarial Imitation Learning}~\cite{ho2016generative} to generate parameters of weather and uses the cross-entropy method for efficient scenario searching.

\subsubsection{Adversarial Policy}

The second type is building a policy model to sequentially control the dynamic objects, which contains the most number of existing works. 
This type is usually formulated as a Reinforcement Learning (RL) problem~\cite{sutton2018reinforcement}, where the AV belongs to the environment and the generator is the agent we can control.
Intuitively, we have much more flexibility under this setting, but the complexity also increases.
Since the AV and objects in scenarios interact step-wise, this problem can be formulated in an RL framework, and there are lots of works using RL methods.
\cite{feng2021intelligent} and \cite{sun2021corner} use Deep Q-Network to generate discrete adversarial traffic scenarios.
\cite{kuutti2020training} uses Advantage Actor Critic (A2C)~\cite{mnih2016asynchronous} to control one surrounding vehicle in the car following scenarios.
\cite{chen2021adversarial} uses Deep Deterministic Policy Gradient (DDPG)~\cite{lillicrap2015continuous} to generate adversarial policy to control surrounding agents to generate lane-changing scenarios.
\cite{wachi2019failure} uses Multi-agent DDPG~\cite{lowe2017multi} to control two surrounding vehicles (which are called Non-player Characters) to attack the ego vehicle. This method also sets auxiliary goals for Non-player characters to avoid generating unrealistic scenarios.
\cite{nonnengart2019crisgen} proposes CriSGen that uses constraint-based optimization and \cite{abeysirigoonawardena2019generating} uses Bayesian Optimization.

A series of works on generating risky and adversarial scenarios in the context of IS also appear in the literature. 
\cite{zhao2015accelerated} uses heuristic approaches to generate dangerous lane change scenarios. 
\cite{7534875} constructs IS distribution to sample dangerous AV lane change scenarios.
\cite{zhao2018accelerated} extends IS approach to dynamic systems to sample dangerous AV car-following scenarios. 
\cite{7989024} uses piece-wise models to design a more expressive IS distribution. 
\cite{huang2018versatile} uses Gaussian Mixture Model (GMM) for IS distribution and further analyzes the efficiency of GMM-based IS distribution for random forest and NN classifiers~\cite{huang2018rare}. 
ReLU-activated NNs are considered in~\cite{arief2020deep} to estimate the dangerous set and compute an IS estimator for a risk upper bound for Gaussian case, with a more general case presented in \cite{ arief2021certifiable}. The Adaptive IS approach is used to construct adversarial environments to accelerate policy evaluation~\cite{xu2021accelerated}. 
Finally, there is a series of works named Adaptive Stress Testing (AST) that explores different ways to generate stress testing scenarios. 
\cite{lee2015adaptive} uses Monte Carlo tree search (MCTS) to search action of testing scenario, but this method does not target autonomous driving systems.
\cite{koren2018adaptive} generates a scenario controlling a pedestrian to cross the road.
\cite{koren2019efficient} improves the last paper by using LSTM to generate initial conditions and actions in each step.
Instead of defining heuristic reward functions, \cite{koren2020adaptive} leverage the Go-Explore framework to find failure cases.
\cite{koren2021finding} extends previous works to high-fidelity simulation and changes the learning algorithm to Proximal Policy Optimization (PPO)~\cite{schulman2017proximal}.



\section{Knowledge-based Generation}
\label{sec:knowledge}

In Section~\ref{sec:data} and Section~\ref{sec:adversarial}, we discussed methods that purely use data or interact with the AV to generate scenarios. 
However, scenarios are constructed in the physical world, therefore, need to satisfy traffic rules and physical laws. 
The samples in the density we estimate or the adversarial examples we generate could easily violate these constraints. 
In addition, domain knowledge also improves the efficiency of the generation.
\update{After traffic accidents happen, we humans analyze the scenario and find the reasons that cause the accidents. Finding the underlying causality, e.g., the view of the sensor is blocked, is important to efficiently generate safety-critical scenarios.
}

Therefore, in this section, we consider methods that incorporate external domain knowledge into the generation process. We will first explore rule-based methods that artificially design the structure and parameters of scenarios. Then, we turn to the learning-based methods that use explicit knowledge to guide the generation. Assume we can obtain certain domain knowledge $\mathcal{K}$ from experts, then we can augment the learning with
\begin{equation}
    \hat{\theta} = \arg\max \mathbb{E}_{x\sim p_{\theta}(x | \mathcal{K})}\left[ g(F(x)) \right]
\end{equation}
where the scenarios are sampled from a conditional distribution $p_{\theta}(x | \mathcal{K})$.
In addition, we can also use $\mathcal{K}$ as constraints to manipulate existing scenarios:
\begin{equation}
    \tilde{x} = \arg\max g(F(x))\ \ s.t.\ \mathcal{K}(x) \geq 0
\end{equation}
where $\mathcal{K}(x) \geq 0$ means the constraint is satisfied.

\subsection{Pre-defined Rules}
\label{sec:knowledge1}

Driving scenarios are very common in daily life thus we humans can easily design scenarios with pre-defined rules and specific conditions to trigger events. 
\cite{rana2021building} uses prior knowledge to design random risk scenarios with equations. The authors also conduct interventional experiments by training RL agents on different scenarios to get comparable results.
\cite{menzel2018scenarios} focuses on functional and logical scenarios, which can be represented by natural language from human experts.
\cite{bagschik2018ontology} views ontologies as knowledge-based systems in the field of AV and proposes a generation of traffic scenes in natural language as a basis for scenario creation.
\update{In \cite{fremont2020formal}, the authors combines formal specification~\cite{lamsweerde2000formal} of scenarios and safety properties to generate test cases from formal simulation.}

There are also a large number of works that build the entire platform with pre-defined scenarios implemented. We categorize them under the knowledge-based generation method and will also discuss more details about these platforms in Section~\ref{sec:tools}.
A 2D platform named {SMARTS} is developed in \cite{zhou2020smarts} containing multiple diverse behavior models using both rule-based and learning-based models.
\cite{li2021metadrive} proposes {MetaDrive}, a 3D simulator that supports different road shapes defined by users or directly imported from existing real-world datasets (e.g. Argoverse~\cite{chang2019argoverse}).
\cite{carlachallenge} is a competition built on top of CARLA~\cite{dosovitskiy2017carla} and~\cite{scenariorunner}, which consists of a large number of pre-defined scenarios.
In \cite{didrive}, the authors build a rule-based scenario casezoo in CARLA~\cite{dosovitskiy2017carla}, which shares some similar scenarios with~\cite{scenariorunner}.
To manage complex traffic scenarios with hundreds of objects, {SUMMIT}~\cite{cai2020summit} is specifically designed for generating massive mixed traffic with an autopilot algorithm.
To explore the causality between vehicles in the scenario, {CausalCity}~\cite{mcduff2021causalcity} is developed for evaluating causal discovery algorithms. It builds an agency mechanism to define high-level behaviors.

\subsection{Knowledge-guided Learning}
\label{sec:knowledge2}

\subsubsection{Knowledge as Condition}
Combining learning methods and domain knowledge is a popular trend in the machine learning area. One of the biggest obstacles is how to define the representation space of knowledge. This representation should be easily integrated into NNs and still has interpretability.
In \cite{shiroshita2020behaviorally}, the authors assume that diversity and high skill are important for scenarios. They use RL methods to generate diverse scenarios and show improvement of the AV trained in their scenarios.
\cite{ding2021semantically} represent the explicit knowledge (e.g., the vehicles should not have overlap, the orientation of vehicles should follow the direction of the lane) as first-order-logic~\cite{smullyan1995first}, which can be embedded into a tree structure. Then they search in the latent space of a VAE model and apply the knowledge to the tree encoder to constraint the searching process. They evaluate their scenarios on point cloud segmentation methods and show that their scenario can cause failure to some models that are not robust.
In \cite{ding2021causalaf} and \cite{ding2022generalizing}, the authors explore the fundamental reason for the safety-critical scenario, which can be represented by causality. This work assumes there is a causal graph that can represent the relationships between objects in a scenario. Then, they propose an autoregressive generative model that can use this graph to increase the efficiency of the generation.

\subsubsection{Constraint Optimization}
Another way of using explicit knowledge is resorting to the constraint optimization framework. We know that safety-critical scenarios are extremely rare in the real-world log, and random augmentation could be inefficient to generate safety-critical scenarios. 
One of the heuristic metrics of risk is the drivable area for the autonomous vehicle. 
\cite{klischat2019generating} and \cite{althoff2018automatic} minimize the drivable area by controlling the surrounding vehicles with an evolutionary method and constraint optimization methods, respectively.
To explore more different kinds of scenarios, \cite{rocklage2017automated} generates the motion of other traffic participants with a backtracking search.
To make the scenarios diverse, \cite{klischat2020scenario} build a pipeline that introduces the road topology from OpenStreetMap~\cite{bennett2010openstreetmap}.
Using the safety-critical scenarios from \cite{klischat2020scenario}, \cite{wang2021commonroad} designs a comprehensive open-source toolbox to train and evaluate RL motion planners for AVs with customized configuration from users.
To obtain a robust trajectory prediction model, \cite{zhang2022adversarial} and \cite{cao2022robust} generate adversarial trajectory by perturbing existing trajectory with feasible constraints.

\section{Dataset and Tools for Scenario Generation}
\label{sec:tools}

In this section, we introduce the tools that are useful for safety-critical scenario generation. We first discuss the scenario datasets, then turn to the traffic simulators. Finally, we review existing platforms that support the function of scenario generation.

\subsection{Scenario Dataset}
\begin{table*}[ht]
    \caption{Comparison of Scenario Datasets. We list existing datasets and different aspects that we are interested in. \checkmark/$\times$ in the weather column means the dataset contains/doesn't contain data under various weather conditions. BEV: bird's-eye view, FPV: first-person view. D: daytime, N: nighttime. H: highway, I: intersection, RA: roundabout, C: campus, U: urban, S: suburban, R: rural.}
    \label{tab:dataset}
    \renewcommand\arraystretch{1.2}
    \centering
    \begin{tabular}{c|c|c|c|c|c|c|c|c|c|c|c|c|c|c}
    \hline
    \multirow{2}{*}{Dataset} & \multirow{2}{*}{\update{Year}} & \multirow{2}{*}{Real} & \multirow{2}{*}{View} & \multicolumn{4}{c|}{Data Sensor} & \multicolumn{3}{c|}{Annotation} & \multicolumn{4}{c}{Traffic Condition} \\
    \cline{5-15}
    & & & & Image & LiDAR & \update{RADAR} & Traj. & 3D & 2D & Lane & Weather & Time & Region & \update{Jam} \\
    \rowcolor{gray!30}
    \hline
    CamVid~\cite{brostow2009semantic} & 2009 & \checkmark  & FPV  & RGB & $\times$ & $\times$ & $\times$  & $\times$  & \checkmark  & \checkmark  & $\times$  & D  & U & $\times$ \\
    \hline
    KITTI~\cite{geiger2013vision} & 2013 & \checkmark  & FPV  & RGB/Stereo & \checkmark & $\times$ & $\times$  & \checkmark  & \checkmark  & \checkmark  & $\times$  & D  & U/R/H & $\times$  \\
    \hline
    \rowcolor{gray!30}
    Cyclists~\cite{li2016new} & 2016 & \checkmark  & FPV  & RGB & $\times$ & $\times$ & $\times$  & $\times$  & \checkmark  & $\times$  & $\times$  & D  & U & $\times$  \\
    \hline
    Cityscapes~\cite{cordts2016cityscapes} & 2016 & \checkmark  & FPV  & RGB/Stereo & $\times$ & $\times$ & \checkmark  & \checkmark  & \checkmark  & $\times$  & $\times$  & D  & U & \checkmark  \\
    \hline
    \rowcolor{gray!30}
    SYNTHIA~\cite{ros2016synthia} & 2016 & $\times$  & FPV  & RGB & $\times$ & $\times$ & $\times$  & \checkmark  & \checkmark  & $\times$  & \checkmark  & D/N  & U & \checkmark  \\
    \hline
    Campus~\cite{robicquet2016learning} & 2016 & \checkmark  & BEV  & RGB & $\times$ & $\times$ & \checkmark  & $\times$  & \checkmark  & \checkmark  & $\times$  & D  & C & $\times$ \\
    \hline
    \rowcolor{gray!30}
    \update{RobotCar~\cite{RobotCarDatasetIJRR}} & 2016 & \checkmark  & FPV  & RGB & \checkmark & $\times$ & \checkmark  & \checkmark  & \checkmark  & $\times$  & \checkmark  & D/N  & U & $\times$ \\
    \hline
    Mapillary~\cite{neuhold2017mapillary} & 2017 & \checkmark  & FPV  & RGB & $\times$ & $\times$ & $\times$  & $\times$  & \checkmark  & \checkmark  & \checkmark  & D/N  & U & $\times$  \\
    \hline
    \rowcolor{gray!30}
    P.F.B.~\cite{richter2017playing} & 2017 & $\times$  & FPV  & RGB & $\times$ & $\times$ & \checkmark  & \checkmark  & \checkmark  & $\times$  & \checkmark  & D/N  & U & $\times$  \\
    \hline
    BDD100K~\cite{yu2020bdd100k} & 2018 & \checkmark  & FPV  & RGB & $\times$ & $\times$ & \checkmark  & $\times$  & \checkmark  & \checkmark  & \checkmark  & D  & U/H  & $\times$ \\
    \hline
    \rowcolor{gray!30}
    HighD~\cite{krajewski2018highd} & 2018 & \checkmark  & BEV  & RGB  & $\times$ & $\times$ & \checkmark  & $\times$  & \checkmark  & $\times$  & $\times$  & D  & H & \checkmark \\
    \hline
    Udacity~\cite{udacity} & 2018 & \checkmark  & FPV  & RGB & $\times$ & $\times$ & $\times$  & $\times$  & \checkmark  & $\times$  & $\times$  & D  & U & $\times$  \\
    \hline
    \rowcolor{gray!30}
    KAIST~\cite{choi2018kaist} & 2018 & \checkmark  & FPV  & RGB/Stereo & \checkmark & $\times$ & \checkmark  & $\times$  & \checkmark  & $\times$  & $\times$  & D/N  & U & $\times$ \\
    \hline
    Argoverse~\cite{Argoverse} & 2019 & \checkmark  & FPV  & RGB/Stereo & \checkmark & $\times$ & \checkmark  & \checkmark  & $\times$  & \checkmark  & \checkmark  & D/N  & U & $\times$  \\
    \hline
    \rowcolor{gray!30}
    TRAF~\cite{chandra2019traphic} & 2019 & \checkmark  & FPV  & RGB & $\times$ & $\times$ & \checkmark  & $\times$  & \checkmark  & $\times$  & \checkmark  & D/N  & U & $\times$  \\
    \hline
    ApolloScape~\cite{huang2019apolloscape} & 2019 & \checkmark  & FPV  & RGB/Stereo & \checkmark & $\times$ & \checkmark  & \checkmark  & \checkmark  & \checkmark  & \checkmark  & D  & U & \checkmark  \\
    \hline
    \rowcolor{gray!30}
    ACFR~\cite{zyner2019acfr} & 2019 & \checkmark & BEV  & RGB & $\times$ & $\times$ & \checkmark  & $\times$  & \checkmark  & $\times$  & $\times$  & D  & RA & $\times$  \\
    \hline
    H3D~\cite{patil2019h3d} & 2019 & \checkmark  & FPV  & RGB & \checkmark & $\times$ & \checkmark  & \checkmark  & $\times$  & $\times$  & $\times$  & D  & U & \checkmark  \\
    \hline
    \rowcolor{gray!30}
    \scriptsize{INTERACTION~\cite{interactiondataset}} & 2019 & \checkmark  & BEV  & RGB & $\times$ & $\times$ & $\times$  & $\times$  & \checkmark  & \checkmark  & $\times$  & D  & I/RA & $\times$ \\
    \hline
    InD~\cite{bock2020ind} & 2020 & \checkmark  & BEV  & RGB & $\times$ & $\times$ & \checkmark  & $\times$  & \checkmark  & $\times$  & $\times$  & D  & I & $\times$ \\
    \hline
    \rowcolor{gray!30}
    RounD~\cite{krajewski2020round} & 2020 & \checkmark  & BEV  & RGB & $\times$ & $\times$ & \checkmark  & $\times$  & \checkmark  & $\times$  & $\times$  & D  & RA & $\times$ \\
    \hline
    nuScenes~\cite{caesar2020nuscenes} & 2020 & \checkmark  & FPV  & RGB & \checkmark & \checkmark & $\times$  & \checkmark  & \checkmark  & $\times$  & \checkmark  & D/N  & U & $\times$  \\
    \hline
    \rowcolor{gray!30}
    Lyft Level 5~\cite{houston2020one} & 2020 & \checkmark  & BEV  & RGB & \checkmark & \checkmark & \checkmark  & $\times$  & \checkmark  & \checkmark  & $\times$  & D  & S & $\times$ \\
    \hline
    Waymo Open~\cite{sun2020scalability} & 2020 & \checkmark  & FPV  & RGB & \checkmark & $\times$ & $\times$  & \checkmark  & \checkmark  & \checkmark  & \checkmark  & D/N  & U/S & $\times$ \\
    \hline
    \rowcolor{gray!30}
    A*3D~\cite{pham20203d} & 2020 & \checkmark  & FPV  & RGB & \checkmark & $\times$ & $\times$  & \checkmark  & \checkmark  & $\times$  & \checkmark  & D/N  & U & \checkmark  \\
    \hline
    \scriptsize{\update{RobotCar Radar~\cite{RadarRobotCarDatasetICRA2020}}} & 2020 & \checkmark  & FPV  & RGB & \checkmark & \checkmark & \checkmark  & $\times$  & $\times$  & $\times$  & \checkmark  & D/N  & U & $\times$  \\
    \hline
    \rowcolor{gray!30}
    Argoverse 2~\cite{wilson2021argoverse} & 2021 & \checkmark  & FPV  & RGB/Stereo & \checkmark & $\times$ & \checkmark  & \checkmark  & $\times$  & \checkmark  & \checkmark  & D/N  & U & $\times$  \\
    \hline
    \update{PandaSet~\cite{xiao2021pandaset}} & 2021 & \checkmark  & FPV  & RGB & \checkmark & $\times$ & \checkmark  & \checkmark  & $\times$  & $\times$  & $\times$  & D/N  & U & $\times$ \\
    \hline
    \rowcolor{gray!30}
    \update{ONCE~\cite{mao2021one}} & 2021 & \checkmark  & FPV  & RGB/Stereo & \checkmark & $\times$ & $\times$  & \checkmark  & \checkmark  & $\times$  & \checkmark  & D/N  & U & $\times$ \\
    \hline
    \end{tabular}
\end{table*}

For modern machine learning methods, datasets are crucial and necessary. Specifically, for the scenario generation task, there are also many published datasets by companies and academic organizations. In Table~\ref{tab:dataset}, we summarize and compare available scenario datasets in various aspects that we are especially interested in.

\subsubsection{Fidelity}

In Table~\ref{tab:dataset}, we include datasets that are either collected from onboard sensors on public roads or synthetic virtual worlds simulated by traffic simulators. Collecting real-world data is very time-consuming, which requires humans to operate vehicles or drones to record data in the real world in a variety of environments. However, these data are more representative of real-world data distribution. Models developed with these realistic data can be directly applied to the real world. Synthetic datasets, on the other hand, are simple to collect. But they heavily rely on the authenticity of traffic simulators, which usually fail to accurately imitate and render real-world data. 

\subsubsection{Collect View}

In addition to different levels of reality, we also present datasets with different views. For example, HighD~\cite{krajewski2018highd}, InD~\cite{bock2020ind}, and RounD~\cite{krajewski2020round} datasets collect data in bird's-eye view (BEV), which is recorded from drone cameras. KITTI~\cite{geiger2013vision} and Argoverse~\cite{wilson2021argoverse} datasets collect data in first-person view (FPV), which is captured by cameras in front of a car. The BEV data happens in a fixed region, therefore it is more useful to analyze the behavior of objects in a fixed background. In contrast, FPV data is more suitable to train AV algorithms that take egocentric information.

\subsubsection{\update{Data Sensor}}

For each dataset, we examine whether it has the following data types: \update{RGB image, stereo image, LiDAR data, Radar, and trajectory}. All datasets included in the table have RGB images since it is a common data type in traffic scenarios. RGB images are usually collected by cameras mounted on vehicles or drones, and they can be utilized for a variety of computer vision tasks such as object detection and image segmentation. Stereo images are captured by stereo cameras and LiDAR data is collected by LiDAR sensors. Both of them provide 3D information that is particularly useful in 3D tasks such as 3D object detection. 
\update{RADAR is also a common sensor that returns similar 3D information as LiDAR but at a cheaper price. It can work in harsher conditions (e.g., rain and storm) due to the longer wavelength than LiDAR's lights.}
The trajectory data is either recorded by sensors such as GPS or converted from object detection results. This type of data is frequently used in trajectory prediction tasks for planning and control purposes.

\subsubsection{Annotation Type}

The availability of various types of annotations is crucial to each dataset, as it determines which tasks the dataset may be used for. We explore three forms of data annotations in Table~\ref{tab:dataset}: 3D object annotations, 2D object annotations, and lane annotations. 3D annotations can be divided into two categories: 3D bounding box annotations and 3D point cloud annotations. A 3D bounding box annotation describes a cube that exactly holds one specific object. 3D point cloud annotations assign point-wise labels to each point in the point cloud, indicating the point's category. Many tasks in autonomous driving, such as 3D object detection and 3D segmentation, require 3D annotations. Similarly, 2D annotations include 2D bounding box annotations and pixel-wise 2D semantic annotations. Lane annotations describe different types of lanes in the data, as well as the boundaries of drivable regions. This lane information can be used to integrate map and traffic rule information into the algorithm of AVs, enabling more efficient decision-making functions.

\subsubsection{Traffic Condition}

Datasets with a limited level of diversity in conditions and situations are skewed to redundant and highly safe scenarios, which leads to the long tail problem~\cite{cui2019class, lee2019learning}. In order to evaluate the performance of AVs in various scenarios, a dataset must include data under a variety of settings. We mainly consider the following \update{four} key aspects: weather, time, \update{region, and traffic density}. The weather conditions change the entire world as well as the style of the data collected by sensors. For example, the nuScenes dataset~\cite{caesar2020nuscenes} includes data from sunny, rainy, and cloudy conditions. As a result, the images have different appearances depending on the weather. The time condition specifies whether the data was obtained during the day or at night. The main distinction between these two scenarios is the lighting. The region denotes the location from which the data is gathered. For example, the INTERACTION dataset~\cite{interactiondataset} is collected at intersections and roundabouts, whereas the KITTI dataset~\cite{geiger2013vision} is collected in urban, rural, and highway settings. 
\update{Finally, traffic density considers the number of objects in traffic scenarios. 
Higher density indicates a traffic jam or congestion, which requires traffic participants to pay more attention to surrounding objects and take action more carefully.}

\subsection{Traffic Simulation Platforms}



\begin{table*}[ht]
    \caption{Comparison of Traffic Simulation}
    \label{tab:simulator}
    \renewcommand\arraystretch{1.2}
    \centering
    \begin{tabular}{c|c|c|c|c|c|c|c|c|c|c}
    \hline
    \multirow{2}{*}{Simulator}  &  \multirow{2}{*}{\update{Year}}  & \multirow{2}{*}{\makecell[c]{Open \\ Source}} & \multirow{2}{*}{\makecell[c]{Realistic \\ Perception}} & \multirow{2}{*}{\makecell[c]{Customized \\ Scenario}} & \multirow{2}{*}{\makecell[c]{Back-end}} & \multicolumn{2}{c|}{Map Source}  & \multicolumn{3}{c}{API Support} \\
    \cline{7-11}
    & & & & & & Real World  & Human Design  & Python  & C++  & ROS  \\
    \hline
    \rowcolor{gray!30}
    TORCE~\cite{wymann2000torcs}                 & 2000 & \checkmark  & \checkmark  & $\times$  & None  & $\times$  & \checkmark  & $\times$  & \checkmark  & $\times$ \\
    \hline
    Webots~\cite{Webots04}                       & 2004 & \checkmark  & \checkmark  & \checkmark  & ODE  & \checkmark  & \checkmark  & \checkmark  & \checkmark  & \checkmark \\  
    \hline
    \rowcolor{gray!30}
    CarRacing~\cite{CarRacing}                   & 2016 & \checkmark  & $\times$  & $\times$  & None  & $\times$  & \checkmark  & \checkmark  & $\times$  & $\times$ \\
    \hline
    CARLA~\cite{dosovitskiy2017carla}            & 2017 & \checkmark  & \checkmark  & \checkmark  & UE4  & $\times$  & \checkmark  & \checkmark  & \checkmark  & \checkmark  \\
    \hline
    \rowcolor{gray!30}
    SimMobilityST~\cite{azevedo2017simmobility}  & 2017 & \checkmark  & $\times$  & \checkmark  & None  & $\times$  & \checkmark  & \checkmark  & $\times$  & $\times$ \\
    \hline
    GTA-V~\cite{richter2017playing}              & 2017 & $\times$  & \checkmark  & \checkmark  & RAGE  & $\times$  & $\times$  & $\times$  & $\times$  & $\times$ \\
    \hline
    \rowcolor{gray!30}
    highway-env~\cite{highway-env}               & 2018 & \checkmark  & $\times$  & \checkmark  & None  & $\times$  & \checkmark  & \checkmark  & $\times$  & $\times$ \\
    \hline
    Deepdrive~\cite{deepdrive}                   & 2018 & \checkmark  & \checkmark  & \checkmark  & UE4  & $\times$  & \checkmark  & \checkmark  & \checkmark  & $\times$ \\
    \hline
    \rowcolor{gray!30}
    esmini~\cite{esmini}                         & 2018 & \checkmark  & \checkmark  & \checkmark  & Unity  & $\times$  & \checkmark  & \checkmark  & \checkmark  & $\times$ \\
    \hline
    AutonoViSim~\cite{best2018autonovi}          & 2018 & $\times$  & \checkmark  & \checkmark  & PhysX  & $\times$  & \checkmark  & $\times$  & $\times$  & $\times$ \\
    \hline
    \rowcolor{gray!30}
    AirSim~\cite{shah2018airsim}                 & 2018 & \checkmark  & \checkmark  & \checkmark  & UE4  & $\times$  & \checkmark  & \checkmark  & \checkmark  & \checkmark \\
    \hline
    SUMO~\cite{lopez2018microscopic}             & 2018 & \checkmark  & $\times$  & \checkmark  & None  & \checkmark  & \checkmark  & \checkmark  & \checkmark  & $\times$ \\  
    \hline
    \rowcolor{gray!30}
    Apollo~\cite{apollosimulation}               & 2018 & \checkmark  & $\times$  & \checkmark  & Unity  & $\times$  & \checkmark  & \checkmark  & \checkmark  & $\times$ \\
    \hline
    Sim4CV~\cite{muller2018sim4cv}               & 2018 & \checkmark  & \checkmark  & \checkmark  & UE4  & $\times$  & \checkmark  &  \checkmark  & \checkmark  & $\times$ \\  
    \hline
    \rowcolor{gray!30}
    SUMMIT~\cite{cai2020summit}                  & 2020 & \checkmark  & \checkmark  & $\times$  & UE4  & \checkmark  & \checkmark  & \checkmark  & $\times$  & \checkmark \\
    \hline
    MultiCarRacing~\cite{SSG2020}                & 2020 & \checkmark  & $\times$  & $\times$  & None  & $\times$  & \checkmark  & \checkmark  & $\times$  & $\times$ \\
    \hline
    \rowcolor{gray!30}
    SMARTS~\cite{zhou2020smarts}                 & 2020 & \checkmark  & $\times$  & \checkmark  & None  & $\times$  & \checkmark  & \checkmark  & $\times$  & $\times$ \\
    \hline
    LGSVL~\cite{rong2020lgsvl}                   & 2020 & \checkmark  & \checkmark  & \checkmark  & Unity  & \checkmark  & \checkmark  & \checkmark  & $\times$  & \checkmark \\
    \hline
    \rowcolor{gray!30}
    CausalCity~\cite{mcduff2021causalcity}       & 2021 & \checkmark  & \checkmark  & \checkmark  & UE4  & $\times$  & \checkmark  & \checkmark  & $\times$  & $\times$ \\
    \hline
    MetaDrive~\cite{li2021metadrive}             & 2021 & \checkmark  & \checkmark  & \checkmark  & Panda3D  & \checkmark  & \checkmark  & \checkmark  & $\times$  & $\times$ \\
    \hline
    \rowcolor{gray!30}
    L2R~\cite{herman2021learn}                   & 2021 & \checkmark  & \checkmark  & \checkmark  & UE4  & \checkmark  & \checkmark  & \checkmark  & $\times$  & $\times$ \\
    \hline
    AutoDRIVE~\cite{samak2021autodrive}          & 2021 & \checkmark  & \checkmark  & \checkmark  & Unity  & $\times$  & \checkmark  & \checkmark  & \checkmark  & \checkmark \\
    \hline
    \end{tabular}
\end{table*}

In Table~\ref{tab:simulator}, we summarize existing traffic simulators and compare them in different aspects that we are particularly interested in. 

\subsubsection{Open Source}

Open-source simulators are easily customized, which enables users to design and evaluate various safety-critical scenarios easily, as these scenarios are typically rare and sophisticated and require a high level of customization. Therefore, whether the simulator is open source becomes one of our primary considerations.
Most of the simulation platforms we list in Table~\ref{tab:simulator} are open source. 
Simulators such as {AirSim}~\cite{shah2018airsim} and {SMARTS}~\cite{zhou2020smarts}, for example, release their source code to the public, making it easier for users to modify and enrich the testing environment.

\subsubsection{Realistic Perception}

\begin{figure}
    \centering
    \includegraphics[width=0.48\textwidth]{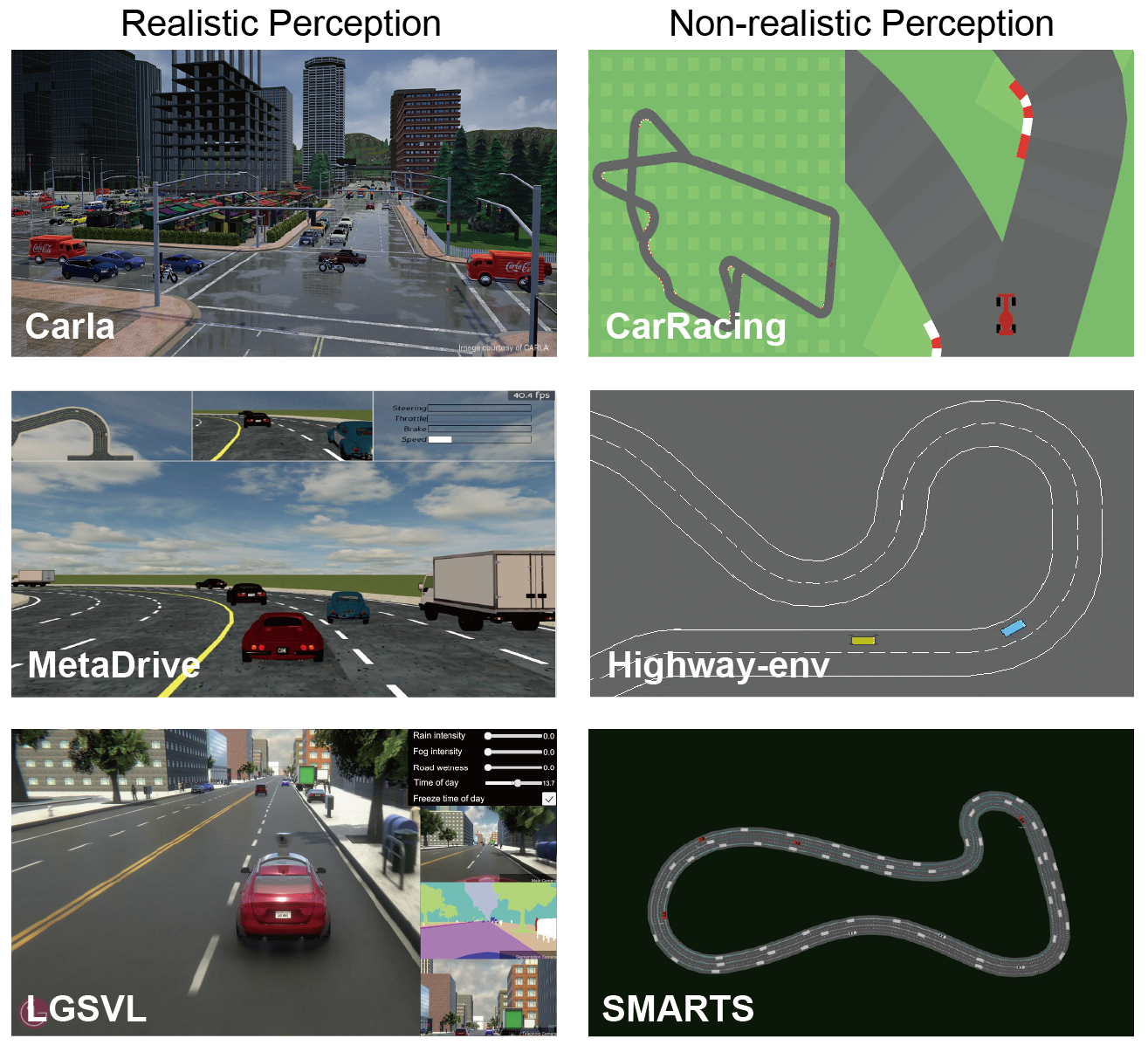}
    \caption{Screenshots of six typical simulators.}
    \label{fig:simulators}
\end{figure}

The realism and fidelity in the virtual simulation have a significant impact on AV algorithms. High-fidelity, photo-realistic simulators provide data that is similar to the physical world, allowing for a more accurate assessment of AV performance in the real world. For example, {SUMMIT}~\cite{cai2020summit} is able to simulate 3D towns with a large number of vehicles, pedestrians, and buildings. Besides, weather conditions can be further simulated, which include controlling the strength of precipitation, cloudiness, fog density, etc. Simple traffic simulators, on the other hand, are incapable of supporting such detailed simulations. However, they are usually light-weighted and easy to use (especially for RL algorithms), where algorithms can be tested quickly without complicated configurations. For example, Highway-env~\cite{highway-env} is a 2D simulator that can be installed using only one command and provide preliminary experimental results in a short amount of time. We provide screenshots of six typical simulators in Figure~\ref{fig:simulators}.

\subsubsection{Customized Scenario}

The freedom to customize scenarios is of great importance since we mainly focus on the generation and evaluation of safety-critical scenarios. The customization of scenarios usually involves the modification of vehicles' positions, speeds, and behaviors. For example, {CARLA}~\cite{dosovitskiy2017carla} offers a systematical way for users to define a scenario, through which users can specify the number of vehicles in a town as well as their own behaviors. Even weather and the surrounding environment can be arbitrarily adjusted to create variations in the visual appearance of the scene. These functions provide higher flexibility for users and allow for more comprehensive testing of the reliability of AVs.

\subsubsection{Back-end Engine}

The simulator engines have a direct impact on the fidelity of the simulated vehicle dynamics and the rendered 3D environment. Most simulation platforms are built upon Unreal Engine 4 (UE4) ~\footnote{https://www.unrealengine.com} or Unity~\footnote{https://www.unity.com}, which are popular and professional game engines. Other less popular engines, such as Panda3D~\cite{goslin2004panda3d}, are also considered. Some non-realistic environments, such as Highway-env~\cite{highway-env}, do not even need back-end engines due to the low requirement of rendering. A simple graphical user interface (GUI) is sufficient for such type of platform.

\subsubsection{Map Source}

Maps are critical components in AV testing systems. The maps in these traffic simulators are either created by humans or based on real-world data. For example, Learn-to-Race (L2R)~\cite{herman2021learn} platform has three racetracks in its racing simulator, all of which are based on real-world racetracks. Human-designed maps can be further classified into two types: rule-based maps and procedurally generated maps. Highway-env~\cite{highway-env} incorporates $11$ rule-based maps, including highways, parking lots, roundabouts, etc., all of which are manually written by humans. MetaDrive~\cite{li2021metadrive} maintains several basic roadblocks and generates numerous maps in a procedural manner by randomly selecting one roadblock at one time.

\subsubsection{API Support}

API support for specific programming languages is crucial in large-scale automated evaluation since it allows users to run batches of scenarios. Popular programming languages like Python and C++ are supported by most simulation platforms. Some simulators like {CARLA}~\cite{dosovitskiy2017carla} and {LGSVL}~\cite{rong2020lgsvl} also provide support for Robot Operating System (ROS)~\footnote{https://www.ros.org}, allowing users to integrate other open-source modules developed by the ROS community.

\subsection{Scenario Design Platform}

There are several user-friendly platforms that support scenario design, which already implements a lot of rule-based scenarios that are normal or safety-critical. We list some popular platforms in this section.

\textbf{CARLA Scenario Runner}~\cite{scenariorunner} provides traffic scenario definitions and an execution engine for CARLA. Scenarios can be defined through a Python interface that allows users to easily describe sophisticated and synchronized maneuvers that involve multiple entities like vehicles, pedestrians, and other traffic participants. It also supports the OpenSCENARIO~\cite{jullien2009openscenario} standard file format for scenario descriptions, making it simple and efficient to incorporate a variety of existing traffic scenarios from the community.

\update{
\textbf{SCENIC}~\cite{fremont2022scenic} defines a language for scenario specification and generation. It describes distributions over scenes and the behaviors of their agents over time. One advantage of SCENIC over other scenario languages is that it combines the concise, readable syntax for spatiotemporal relationships with the ability to impose hard and soft constraints over the scenario.
}

\update{\textbf{SafeBench}~\cite{xu2022safebench} is an open-source platform focusing on systematically evaluating the safety and robustness of autonomous driving algorithms based on diverse testing scenarios and comprehensive evaluation metrics.
The platform integrates eight types of safety-critical scenarios and incorporates four generation algorithms.
Users can also design their own traffic scenarios and scenario generation algorithms following the instructions.
SafeBench also provides several RL-based autonomous driving algorithms with pre-trained RL model weights.
Users can easily test and improve the generated scenarios based on feedback from diverse autonomous driving algorithms.
}

\textbf{DI-Drive Casezoo}~\cite{didrive} consists of a set of scenarios used to train and evaluate diving policy in a simulator. Similar to CARLA Scenario Runner~\cite{scenariorunner}, DI-drive Casezoo has a routing scenario and multiple single scenarios that can be triggered along the route. There are $18$ route scenarios and $8$ types of single scenarios that can be triggered depending on the route definition.
Route scenario is defined in an XML file with its corresponding scenarios. Trigger locations along the route are defined in JSON files. A single scenario is defined in a Python file, describing the behaviors of traffic participants. 

\textbf{SUMO NETEDIT}~\cite{sumo_netedit} is a graphical scenario editor which can be used to create traffic networks from scratch and to modify all aspects of existing networks, including basic network elements (junctions, edges, and lanes), advanced network elements (e.g., traffic lights), and additional infrastructure (e.g., bus stops). This tool is specifically designed for SUMO~\cite{lopez2018microscopic}, which mainly generates large-scale traffic conditions without high-fidelity rendering.

\textbf{SMARTS Scenario Studio}~\cite{smarts_sstudio} is a scenario design tool in the SMARTS~\cite{zhou2020smarts} platform that supports flexible and expressive scenario specification. Scenario definitions are written in the Domain Specific Language, which describes the traffic environment, such as traffic vehicles, routes, and agent missions. Scenario Studio also supports configuration files from SUMO’s NETEDIT~\cite{sumo_netedit}. Maps edited by NETEDIT~\cite{sumo_netedit} can be easily included and reused in Scenario Studio, which enriches the training and testing environments in the SMARTS platform.

\textbf{CommonRoad}~\cite{wang2021commonroad} is a simulator and an open-source toolbox to train and evaluate RL-based motion planners for AVs. Scenario configurations are written in XML files. Users can read, modify, visualize, and store their own traffic scenarios using the Python API provided by CommonRoad. In addition, CommonRoad also supports more scenario specifications, such as Lanelet2~\cite{poggenhans2018lanelet2} and OpenSCENARIO~\cite{jullien2009openscenario}.


\section{Challenges and Potential Solutions}
\label{sec:directions}

\begin{figure}
    \centering
    \includegraphics[width=0.4\textwidth]{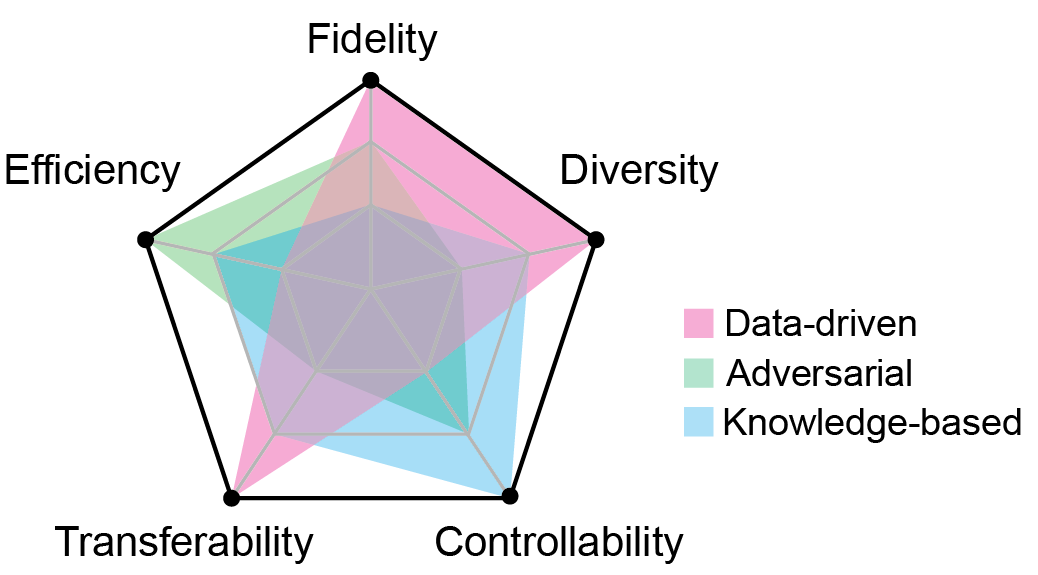}
    \caption{Comparison between three types of generation under the five challenges.}
    \label{fig:challenge}
\end{figure}

In previous sections, we discussed the safety-critical scenario generation methods from three different views: data-driven, adversarial, and knowledge-based. We notice that generating safety-critical scenarios could be a hard problem due to the many constraints and required properties. In this section, we identify five main challenges that cover the difficulty of the generating process, as shown in Figure~\ref{fig:challenge}. Digging into the details of these challenges also helps us discover potential directions that can improve existing algorithms.
The five challenges are as the following:
\begin{itemize}
    \item \textbf{Fidelity.} Our ultimate goal is to develop safe AVs that can run in the real world. Therefore, it is useless to make AVs pass difficult but unrealistic scenarios. We need to ensure that generated scenarios have the chance to happen in real traffic situations.
    \item \textbf{Efficiency.} Safety-critical scenarios are extremely rare in the real world. The generation needs to consider the efficiency and increase the density of the scenarios we are interested in.
    \item \textbf{Diversity.} Safety-critical scenarios are also diverse. The generation algorithm should be able to discover and generate as many different safety-critical scenarios as possible.
    \item \textbf{Transferability.} Scenarios are dynamic due to the interaction between the AVs and their surrounding objects. The scenarios we generate should be variable for different AVs rather than targeting one specific AV.
    \item \textbf{Controllability.} In most times, we want to reproduce or repeat specific scenarios rather than random ones. The generative model should be able to follow instructions or conditions to generate corresponding scenarios.
\end{itemize}
We will discuss more details from the above five perspectives in the following sections, and we will show that the combination of the previous three types of generation methods could be very promising ways to solve those challenges.

\subsection{Fidelity}

The generation algorithm can create infinite scenarios, but not all of them are able to happen in the real world. 
Particularly, under the adversarial generation framework, the searched scenarios are likely to violate the basic traffic rules.
The intuitive way to avoid this problem is adding constraints during the generation, but sometimes the constraints are not easy to define.
Another promising direction is combining real-world data and adversarial generation, where the real-world data can be used as a prior distribution or constraints. Metrics such as Kullback–Leibler divergence~\cite{kullback1951information} and Wasserstein distance~\cite{kantorovich1960mathematical} can be used to minimize the gap between the generated and real-world scenarios.

The fidelity of the scenario is also reflected by the high-dimensional sensing data, which usually requires powerful DGMs to generate. The state-of-the-art methods mainly focus on the generation of static images such as faces. Recently, Neural Radiance Fields (NeRF)~\cite{mildenhall2020nerf} is popular for visual scene generation. This method uses NNs to learn the ray-casting functions and then output different views of a scene. Extending this method to large-scale traffic scenario generation is also an interesting direction, which has been explored in Block-NeRF~\cite{tancik2022block}. Block-NeRF builds a large-scale traffic scene from pure image data. 

\subsection{Efficiency}

Due to the black-box property of most autonomous systems, it is inefficient to generate adversarial examples without accessing the inner information of the systems. In the adversarial attack area, methods with surrogate models~\cite{papernot2017practical} or gradient estimation methods~\cite{guo2019simple} are utilized to tackle this problem. They either learn a differentiable surrogate model to imitate the original autonomous systems or query the system to estimate the approximate direction gradient.  

It is also noticed that uniform sampling from collected data is quite inefficient because of the rareness of safety-critical scenarios. 
Therefore, previous methods propose to use IS, which focuses on the region of the distribution that we are interested in. However, it is difficult to extend IS methods to high-dimensional cases.
In addition, even for the adversarial generation methods, the black-box property of victim AVs is still the biggest obstacle. Without access to the internal information of failure, the generation algorithms are unable to update their scenarios efficiently.

one potential solution to this problem is leveraging the symbolic reasoning~\cite{mao2019neuro} and causal discovery~\cite{spirtes2016causal} techniques. 
Instead of performing the optimization only in large numerical space, using symbolic representation helps the generative model to reason about the elements that make the scenario safety-critical.
Causal discovery methods can uncover the underlying causality behind safety-critical scenarios, finding the mechanisms that cause the risk.

\subsection{Diversity}

Safety-critical scenarios are rare but also diverse. Most of the current generative methods focus on searching for the best scenario that satisfies the requirements but ignores diversity. To comprehensively evaluate the performance of AV, we need a large range of scenarios. It is easy to fall into the risk of over-fitting if the testing scenarios are very similar.
To increase the diversity of the scenario, there are generally two directions. One is from the optimization perspective, where sampled-based methods, such as the evolutionary method or Bayesian Optimization can be used to get feasible solutions from multiple modes.
The other direction is applying regularization to the density estimation model or building multi-modal distribution (e.g., Gaussian Mixture Model) to represent the scenario.


\subsection{Transferability}

In the safety and robustness areas, the adversarial attack is considered a common way to generate risky examples. The transferability, which means the generated samples are also applicable to other algorithms, is a crucial factor in evaluating the generation method.
For example, the pixel-level adversarial attack only works for the target victim.
It is believed that the attack should happen at group level~\cite{xu2018structured} or semantic level~\cite{xiao2018spatially} to achieve better transferability. 
One example in the safety-critical scenario is controlling one surrounding vehicle (SV) to hit the AV step by step. After the policy of the SV is trained, we change the target AV. The new AV shows a very different behavior and follows a route that the SV has never seen before. It is most likely that this scenario is not risky for the new AV.
In this example, we should let the SV learn at a higher level, where it contains the semantic meaning of risk.
The SV can suddenly show up behind another vehicle, which leaves the AV a very short time to react. 
Essentially, we need to build a hierarchical scenario where the high level makes a plan for the risky scenario, and the low level executes that plan with control commands.

\subsection{Controllability}

Controllability is useful in two cases. One is that we want to repeat one specific scenario with several parameters fixed, and the other is generating different scenarios with similar settings. 
For example, we want to test the performance in a highway environment, and then we want to generate vehicles that approach the AV from different directions.
Conditional generative models~\cite{mirza2014conditional, sohn2015learning} are widely used to generate controllable samples, which learn a joint distribution of the condition and the data. 
Sometimes, the condition could be simple as numerical values or as complex as natural languages. 
The challenge is that current NN-based models have poor generalization, therefore, fail when the given unseen conditions during the generating. 
One promising direction could be to increase the generalizability under such a zero-shot setting.

\subsection{Extension to Other Applications}

Autonomous driving is essentially an application of mobile robots. To make the intelligent hardware widely used in the real world, other types of applications should also be evaluated on safety-critical scenarios -- for example, household robots and manipulation tasks.
However, the generation of indoor scenarios could be much more difficult. The traffic scenario basically consists of dynamic objects on a 2D surface, but the indoor scenarios contain a large number of objects interacting in a 3D space. The relations between these objects are also diverse and complex. The direction of how to extend scenario generation methods to these applications is still challenging.

\section{Conclusion and Discussion}
\label{sec:conclusion}

In this survey, we review existing safety-critical scenario generation methods and categorize them into three types. We also review the simulation platforms and datasets that can be used for scenario generation.
Most importantly, we identify five challenges for this topic and point out potential directions to tackle these challenges. 
In the end, we summarize the important message that the readers can take away from this survey, including why is safety-critical scenario generation important, how to select generation algorithms from so many existing methods, and what are future directions to improve existing algorithms.

\subsection{Importance of This Topic}

Most existing driving systems are still evaluated on naturalistic scenarios collected from daily life or human-designed scenarios. Even if they can autonomously drive more than thousands of miles without disengagement, we are still not sure about their safety and robustness, e.g., whether the AV can keep safe when the surrounding vehicles behave aggressively or whether the AV can successfully stop when a pedestrian suddenly runs out from behind an object. 
The generation methods accelerate the development and evaluation of driving systems by creating diverse and realistic safety-critical scenarios. 

\subsection{How to Select Generation Algorithms}

Although we have categorized existing methods with a taxonomy, it is still not clear how to select a specific algorithm according to different situations. We summarized three combinations to help make choices in general cases.
\begin{itemize}
    \item \textbf{Data-driven generation + Adversarial generation}. 
    If the goal is to broadly evaluate your system under diverse scenarios to discover the weakness, combining multi-modal density models and adversarial training is preferred. Randomly sample from the generative models and search the safety-critical scenarios with adversarial generation.
    \item \textbf{Data-driven generation + Knowledge-based generation}. 
    When there are specific requirements that can be converted into constraints, using constraint optimization to manipulate existing scenarios (e.g., trajectories) is preferred. In that case, the scenarios will concentrate on one single cluster with low diversity, which is helpful if the target is to test the driving system on specific scenarios.
    \item \textbf{Adversarial generation + Knowledge-based generation}
    If we already have rules that can design safety-critical scenarios but also want to increase the diversity of generated scenarios by automatically learning the parameters, combining the adversarial generation and rule-based method is proffered. 
\end{itemize}

\subsection{What are Future Directions}

\textbf{Properly integrate inductive bias.} 
We want to emphasize that driving scenarios are created by the relations of objects and physical laws rather than being designed by the human mind. 
Purely relying on NNs or optimization methods is not the ultimate solution for generating realistic and critical scenarios. Instead, the model should leverage as much external knowledge and rules as possible to make the generated scenarios interpretable and satisfy the goal.
For instance, the representation of the scenario is crucial and usually determines the quality of the generation. A representation that naturally embeds rules and laws within a structure could be easily optimized by considering the complex relationship between objects.
\update{In addition, correctly injecting the distribution of real-world data is important to ensure the reality of generated scenarios. Using Offline RL~\cite{prudencio2022survey} and imitation learning could be a potential direction to achieve this goal.  
}

\textbf{Use scenarios to increase robustness and safety.}   
Another aspect that is worth investigating is how to effectively use the generated scenarios to improve robustness and safety. The most intuitive way is training the autonomous system against generated safety-critical scenarios under the adversarial training framework. However, the adversarial scenarios usually represent the worst cases, therefore, learning to a robust yet conservative system. It is not easy to select the difficulty and category of scenarios due to the problems of imbalanced data and over-fitting. 
These problems bridge the topic discussed in this survey to other areas such as robust optimization~\cite{beyer2007robust} and distributional robust optimization~\cite{rahimian2019distributionally}, which have broadly literature to be explored.

\textbf{Use scenarios to improve generalization.}
Besides increasing the robustness, the generated scenarios could also be used to improve the generalization of AVs. For instance, gradually training AVs with increasing risk levels under curriculum learning~\cite{soviany2021curriculum} framework may help systems easily generalize to more types of safety-critical scenarios. One recent survey~\cite{kirk2021survey} that investigates the generalization problem in RL emphasizes the importance of environment generation in increasing the similarity between training and testing domains. This direction extends the scenario generation from safety to broader views that require goal-conditioned environment generation.

\section*{Acknowledgement}
The authors gratefully acknowledge the support from the National Science Foundation (under grants CNS:CAREER-2047454).

\bibliographystyle{IEEEtran}
\bibliography{reference}

\end{document}